# REDUCING RISK FOR ASSISTIVE REINFORCEMENT LEARNING POLICIES WITH DIFFUSION MODELS

## A. TYTARENKO

**Abstract.** Care-giving and assistive robotics, driven by advancements in AI, offer promising solutions to meet the growing demand for care, particularly in the context of increasing numbers of individuals requiring assistance. This creates a pressing need for efficient and safe assistive devices, particularly in light of heightened demand due to war-related injuries. While cost has been a barrier to accessibility, technological progress is able to democratize these solutions. Safety remains a paramount concern, especially given the intricate interactions between assistive robots and humans. This study explores the application of reinforcement learning (RL) and imitation learning, in improving policy design for assistive robots. The proposed approach makes the risky policies safer without additional environmental interactions. Through experimentation using simulated environments, the enhancement of the conventional RL approaches in tasks related to assistive robotics is demonstrated.

## INTRODUCTION

Care-giving and assistive robotics are some of the most promising potential applications for AI systems. For decades already, it has been an active research field. This is certainly unsurprising, given the growing number of people who need care, which at some point may not be difficult to sattisfy in some countries [1, 2].

Moreover, given the circumstances of war actions in Ukraine, the demand will constantly grow. Tens of thousands of people will require rehabilitation, and some of them will require physical assistance for very long periods. To satisfy that demand, some amount of automation is certainly necessary. Although given the high costs of assistive devices, not everybody can afford them, the technological progress and drastic simplification of development and hardware requirements will help to democratize them and make them affordable.

One of the biggest concerns of that progress is safety [3]. At the moment, most devices employ sophisticated manually-designed policies and mechanisms to ensure robustness and safety. Since assistive robots interact with humans, it is desirable to reduce the risk, or in other words, improve the success rate..

One way of automating the policy design process is machine learning. For instance, reinforcement learning (RL) allows for policy learning from data of interaction with humans, which are difficult to rigorously model and predict [3, 4]. Manually designed policies have difficulties with such cases, as it's difficult to make them robust to the modeling errors.

This problem is amplified when humans can demonstrate only limited cooperation, such as in the case of people with disabilities. Cheap robotic arms (which are more affordable) are also difficult to rigorously model, which makes vendors choose expensive hardware instead.

RL already has been applied to tasks that involve difficult-to-model physical tasks, such as ziplock bag manipulation [5] or cable manipulation [6]

However, RL policies often require millions of time steps for full convergence, and while the explored good trajectories are produced much earlier, it takes a while for the policies to stop breaking them for the sake of exploration.

In this paper, a way of not taking the burden of training an algorithm until full convergence, but rather collecting those first successful trajectories and using them to fit a robust policy is considered. This allows for reducing the risk of fairly non-robust policies without any additional interactions with the environment. It is demonstrated that this approach outperforms the bare model-free RL method in the tasks of assistive robotics, simulated using Assistive Gym [7].

**PRELIMINARIES**

**Markov Decision Process** (MDP) $M$ is a tuple $(S, A, r, T)$, where $S$ - state space, $A$ - action space, $r: S \times A \to R$ - reward function, and $T = P(s_{t+1} | s_t, a_t)$ - probability that an environment will transition to the state $s_{t+1}$ given that the current state is $s_t$ and the action taken $a_t$

**Reinforcement Learning** algorithm takes MDP $M$ and searches for a policy $\pi$, that maximizes the discounted return objective:

$$\pi^* = \arg\max_\pi \mathrm{E}_{\tau \sim p(\cdot)} \sum_{t=0}^{\infty} \gamma r(s_t, a_t)$$

Here $\tau$ is a trajectory $(x_0, a_0, x_1, a_1, ... x_T)$ usually sampled by applying a policy $\pi$.

Model-free reinforcement learning algorithms are considered, namely Actor-Critic, which learns a value function $v(s)$ and a policy function $\pi(a|s)$. The latter is often minimized using the former for the advantage estimation. There are multiple instances of the Actor-Critic algorithm. For instance, Proximal Policy Optimization [8] or A3C [9].

Also diffusion models for policy fitting are considered. Namely, Denoising Diffusion Probabilistic Models (DDPM), which are generative models based on Stochastic Langevin Dynamics [10].

The idea is to fit a noise-predicting network $\varepsilon_\theta$, that predicts a gradient $\nabla E(x)$. This gradient is computed and applied repeatedly

$$x' = x - \gamma \nabla E(x)$$

to recover an input $x_0$ from its noised version $x_K$. These models are trained on a set of inputs and then are used to generate novel inputs from pure noise. This approach to imitation learning is a focus of [11]

For assistive robotics, Assistive Gym [7] is used. It is modified it to be more suitable for using it with diffusion policies. See the Experimental Validation section for details.

**METHOD**

First, a model-free reinforcement learning algorithm is employed to discover successful trajectories. Proximal Policy Optimization (PPO) algorithm is chosed in this paper, as it is widely used and is very easy to apply. That algorithm collects the trajectories by interacting with an environment and fits its policy function by minimizing the following loss.

$$L_{PPO}(\phi) = \mathrm{E}_{\tau \sim p_\phi(\cdot)} \min(r_t A_t, clip(r_t, 1-\varepsilon_t, 1+\varepsilon_t) A_t)$$

Specifically, a fully connected network with ReLU activation functions is used to approximate the policy/value functions. This decision was made because a low dimensional state is being observed instead of images.

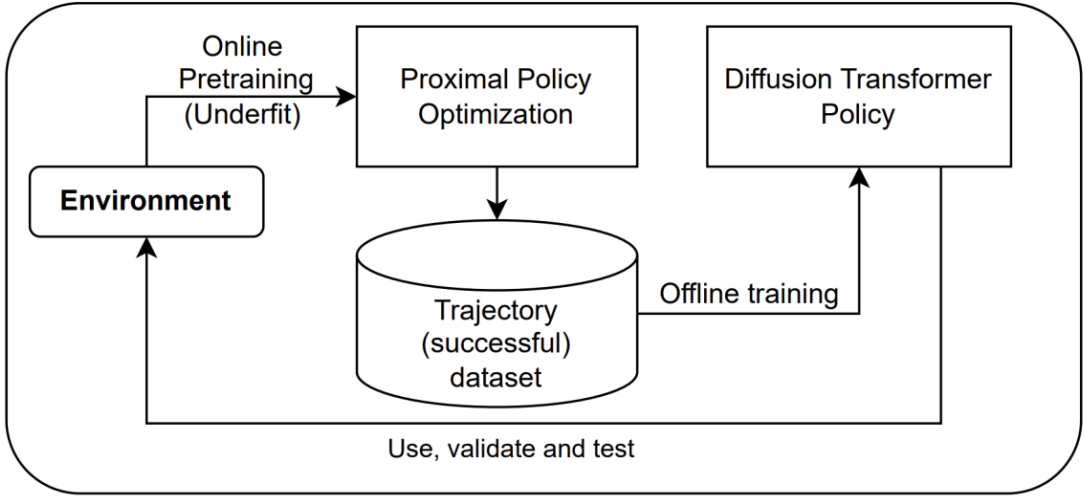

Fig 1. A diagram schematic of a proposed algorithm. First, a baseline online RL policy is pretrained, then successful trajectories are sampled, and a diffusion policies algorithm is fine-tuned on those in an offline manner.

A PPO is trained as a baseline. The problem with PPO is that it requires a lot of samples, which is often too expensive (in computing) or dangerous (when applied in the real-world setting). Therefore, it is trained for a fixed number of time steps, stopping it often way before the full convergence.

When applied to assistive robotics tasks, PPO produces high-risk policies, as evidenced by experiments (see EXPERIMENTAL VALIDATION section).

To reduce the risk, successful trajectories produced by high-risk PPO-based policies are selected and imitation learning techniques based on diffusion models are applied.

$r_s(s_i)$ is defined as a new reward function, which may be different from the original and is binary, meaning it is either 1 or 0. For instance, in the Assistive Feeding task, success is defined like a predicate "The food on a spoon is safely placed in the mouth of a human within 10 seconds". This sparse reward is given right before the episode's termination. For brevity, denote

$$r_s(\tau) = \sum_{s_i \in \tau} r_s(s_i)$$

Also, define $\pi_\theta(a|s, success)$ as a policy conditioned on $(a, s)$ being a part of the successful trajectory. This approach is inspired by control as inference problem statement [12]

Suppose a PPO policy $\pi_\theta$ is trained. If one takes

$$p_\phi(\tau | success) = p(s_0) \prod_{i=1}^{T} \pi_\phi(a_i | s_i, success) p(s_i | s_{i-1}, a_{i-1})$$

and fits the Diffusion Policy on $(s, a) \sim p_\phi(\tau | success)$, they'd arrive to a much more robust policy, given the assumption that the set of successful trajectories is enough to cover the stochasticity and uncertainty of the environment. For the robotics tasks considered, this assumption tends to be true in practice.

The diffusion policy is approximated using a diffusion transformer architecture, proposed in [11]. It is simplified a bit for the environments with shorter trajectories. Also, it has been found that U-Net-based architectures give almost the same results, so it is not include them in this study.

**Algorithm**

Let's summarize an algorithm described earlier.

1. Train a PPO policy for $T_{PPO}$ time steps.

2. Sample a dataset of trajectories using a pre-trained PPO policy.

$$D = \{\tau_i, i = 1,..., N | \tau_i \sim p_{PPO}(\tau)\}$$

3. Sample a dataset of successful trajectories.

$$D_{succ} = \{\tau | \tau \in D, r_s(\tau) = 1\}$$

4. Initialize the weights of a neural net $\theta$

5. Repeat for $M$ epochs:

   (a). Retrieve a batch of trajectories $B$ from the dataset

   $$B = \{(s_k, a_k, ..., s_{k+T}, a_{k+T})_i, i = 1, ..., N_B\}$$

   (b). Following [11] set $A = (a_k, ..., a_{k+T})$ and $O = (s_k, ..., s_{k+T})$, and minimize the DDPM training loss:

   $$L_{DDPM}(B) = MSE(\varepsilon^k, \varepsilon_\theta(O_t, A_t + \varepsilon^k, k))$$

   $$\theta' = \theta - \gamma \nabla L_{DDPM}(B)$$

   (c). Update neural networks' parameters:

   $$\theta' \leftarrow \theta;$$

Here:

- $T_{PPO}$ - num timesteps for the PPO to be trained on.

- $N$ - number of all samples from the PPO policy

- $N_B$ - batch size for the diffusion model,

- $\gamma$ - learning rate

- $T$ - Diffusion Policy horizon. Can also be different for actions and observations. Actions of this horizon are predicted conditioned on observations.

For a neural network architecture, the Diffusion Transformer [11] architecture is used.

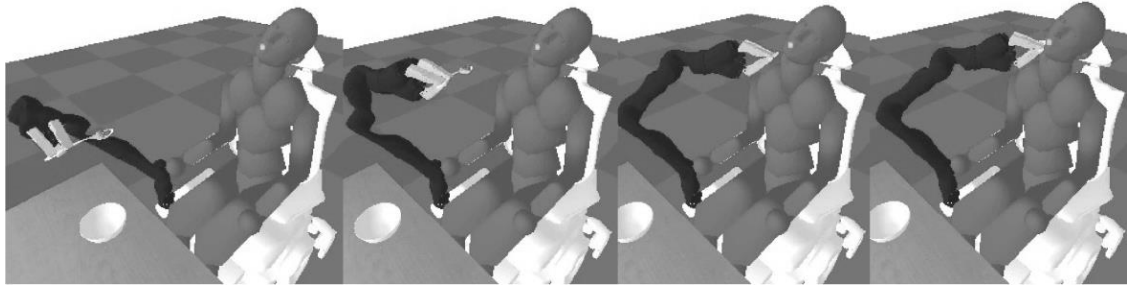

Fig 2. A trajectory produced by the policy trained using the proposed method. The task - Assistive Feeding..

**EXPERIMENTAL VALIDATION**

For experimental validation, Assistive Gym [7] simulation benchmark is used. The observation space usually contains a low dimensional arm state, human head state, and some other task- or tool-related statistics. All the tasks are done using the simulated Jaco robot arm. The proposed method is evaluated on the following tasks:

1. **Assistive Feeding**. A task where a robotic arm uses a spoon full of food to feed a person. The task is considered successful if 75% of food is in the person's mouth. The resulting trajectory is depicted in Figure 2.
2. **Assistive Drinking**. A task where a robotic arm uses a cup full of water to assist a person with drinking. The task is considered successful if 75% of water is in the person's mouth.
3. **Assistive Bed Bathing**. A task where a robotic arm uses a sponge to wash a person. The task is considered successful if the necessary spots of a person's surface are touched with a sponge.
4. **Assistive Arm Manipulation**. A task where a robotic arm is used to reposition a person's arm. The task is considered successful, if the arm is successfully repositioned.

First, a PPO baseline is trained until it is sufficient for a policy to produce successful trajectories. In the study, 1000 successful trajectories have been collected for each task. It has been found, that for many tasks, te number of trajectories less than 300 degrades the performance of the method.

After that, sample successful trajectories are sampled from the PPO policy, as described in Algorithm 1, and a diffusion policy is fit on those.

The results are given in Table 1.

|                  | Success (%)  |                       |
|------------------|--------------|-----------------------|
| Task             | PPO risky    | Fine-tuned (proposed) |
| Arm Manipulation | 19%          | **71%**               |
| Bed Bathing      | 2%           | **12%**               |
| Drinking         | 10%          | **56%**               |
| Feeding          | 33%          | **86%**               |

**Table 1**. Results of fine-tuning the PPO policies with the proposed method. Arm Manipulation and Drinking PPO is trained for 1 million steps. Bed Bathing PPO is trained for 2 million steps (to get any reasonable policies). Feeding PPO is trained for 400k steps. Please note, in order to improve baseline and diffusion performance, episodes are terminated if success is achieved. Originally, termination only occurred when the number of steps exceeded a limit. It's been found that this affects the method's performance.

As one may observe, the resulting policy outperforms the underfit ones, but also performs as good or better than the long-trained PPO policy. Remarkably, this is achieved without any additional environment steps, just using the offline data. One may also sample those trajectories during the training of a baseline, thus removing the need to sample them afterward.

The results that good could be explained by the properties of the modern imitation learning techniques, such as diffusion policy, used in this paper. It has been observed, that these methods demonstrate interesting out-of-distribution generalizations [5].

It is also apparent, that the Bed Bathing benchmark though is improved, but still is not beaten. It is hypothesized, that this is due to a low diversity of the collected trajectories, insufficient to cover the entire distribution.

In addition to generalization, it is hypothesized, that since the PPO itself balances exploration and exploitation, this may prevent it from fast convergence on successful trajectories, continuously trying to look for other modalities.

Another interesting observation is that when the PPO baseline is trained until full convergence using up to several million steps, it usually gets on-par performance with the fine-tuned version. Although, on the Feeding benchmark fully converged PPO got an 87% success rate, while its fine-tuned version achieved 98%.

But even without that, it's still a much worse result than the fine-tuned approach, since it requires additional millions of time steps.

## CONCLUSION

In this paper, a novel approach to reduce risk in assistive reinforcement learning policies using diffusion models is proposed. The proposed method leverages the strengths of both model-free reinforcement learning and imitation learning techniques based on diffusion models to improve policy robustness without additional interactions with the environment.

The effectiveness of the proposed approach is demonstrated through experimental validation on various assistive robotics tasks simulated using Assistive Gym. By fine-tuning policies obtained from a baseline PPO algorithm with offline data, significant improvements in success rates are achieved across different tasks. Importantly, the method outperformed risky policies generated directly by PPO.

The results indicate the potential of diffusion-based imitation learning techniques in enhancing the safety and reliability of assistive robotics systems.

Future work could explore additional refinements to the diffusion-based policy fitting process and include re-exploration iteration for diffusion policies, to make the process iteratively switch between fine-tuning and exploration.